\documentclass{article}
\usepackage{iclr2027_conference,times}
\iclrfinalcopy 
\usepackage[utf8]{inputenc}
\usepackage[T1]{fontenc}
\usepackage{url,booktabs,array}
\usepackage{xurl} 
\usepackage{amsmath,amsfonts,amssymb,amsthm}
\usepackage{algorithm,algpseudocode}
\usepackage{microtype,graphicx}
\usepackage{hyperref}
\newtheorem{proposition}{Proposition}
\theoremstyle{remark}

\newcommand{\best}[1]{\textbf{#1}}
\newcommand{\method}{CoHARL}

\newcommand{\paperfigure}[3]{%
  \IfFileExists{#1}{\includegraphics[width=\linewidth]{#1}}{%
    \fbox{\parbox[c][#2][c]{\dimexpr\linewidth-2\fboxsep-2\fboxrule\relax}{%
      \centering\textbf{Figure placeholder}\par\medskip #3}}}}
\title{\method{}: Co-Evolving Agent Policies and\\Executable Harnesses with Reinforcement Learning}
\author{%
\textbf{Jianwen Chen$^{1}$, Peng Xia$^{1}$, Kaide Zeng$^{1}$, Siwei Han$^{1}$}\\[0.25em]
\textbf{Zhen-Yu Zhang$^{2}$, Charles Fleming$^{3}$, Xinyu Ye$^{1}$, Huaxiu Yao$^{1}$}\\[0.55em]
$^{1}$University of North Carolina at Chapel Hill\\[0.15em]
$^{2}$RIKEN \qquad $^{3}$Cisco}
\makeatletter
\renewcommand{\maketitle}{%
  \par\begingroup\centering
  \hrule height 2pt
  \vspace{12pt}
  {\fontsize{17.28}{20.5}\selectfont\bfseries\@title\par}
  \vspace{12pt}
  \hrule height 0.5pt
  \vspace{17pt}
  {\fontsize{11}{13}\selectfont\normalfont\@author\par}
  \vspace{19pt}
  \endgroup}
\makeatother
\renewenvironment{abstract}{%
  \centerline{\normalsize\bfseries Abstract}\vspace{0.6em}\begin{quote}%
}{\end{quote}\vspace{1ex}}
\hypersetup{hidelinks,pdfauthor={Jianwen Chen, Peng Xia, Kaide Zeng, Siwei Han, Zhen-Yu Zhang, Charles Fleming, Xinyu Ye, Huaxiu Yao},pdftitle={\method{}: Co-Evolving Agent Policies and Executable Harnesses with Reinforcement Learning}}

\begin{document}

\maketitle
\fancyhead{}
\renewcommand{\headrulewidth}{0pt}

\begin{abstract}
Large language model (LLM) agents interact with their environments
through executable harnesses that govern tool use and feedback.
Optimizing a harness for a fixed policy overlooks how its behavior
changes during learning, while training a policy under a fixed harness
leaves the interface that generates its training experience unchanged.
This motivates adapting the harness throughout policy training.
We introduce \method{}, a population-based reinforcement learning framework
that co-evolves a shared policy and a population of executable harnesses.
The policy learns from rollouts across the population, and a proposer
uses its execution traces to revise harness code during training.
\method{} achieves a 27.6\% mean resolved rate on SWE-rebench, improving
over harness-only search by 6.4 percentage points. Experiments on
SWE-bench Verified and Terminal-Bench~2.0 show gains in software
engineering and terminal-based tasks. Under matched training budgets,
\method{} also outperforms searching for a harness once before RL and
alternating policy updates with revisions to a single active harness.
\end{abstract}

\section{Introduction}
\label{sec:intro}
LLM agents solve software-engineering tasks by interacting with
repositories, tools, and execution environments~\citep{weng2026harness,sweagent,openhands,karten2026continual}.
Their behavior depends on both model weights and the harness: the
program that handles prompts, action parsing, tool responses, and
episode control. The same program controls inference-time execution
and the collection of training trajectories.

Agent failures can originate in either the execution code or the policy.
If an action parser rejects a valid tool call, or a tool-output handler
discards feedback needed for the next decision, updating model weights
does not repair that code. Policy-only reinforcement learning
(RL)~\citep{swerl,deepswe} may teach the agent to work around the defect,
but leaves its cause in the harness unchanged. Conversely, even with
correct tool execution, a policy may choose an uninformative command or
misinterpret a test failure. Harness-only search~\citep{metaharness,selfharness,lin2026agentic}
can add guidance and safeguards, but does not update the model's
parameters to learn from these mistakes.

\begin{figure}[t]
\centering
\includegraphics[width=\linewidth]{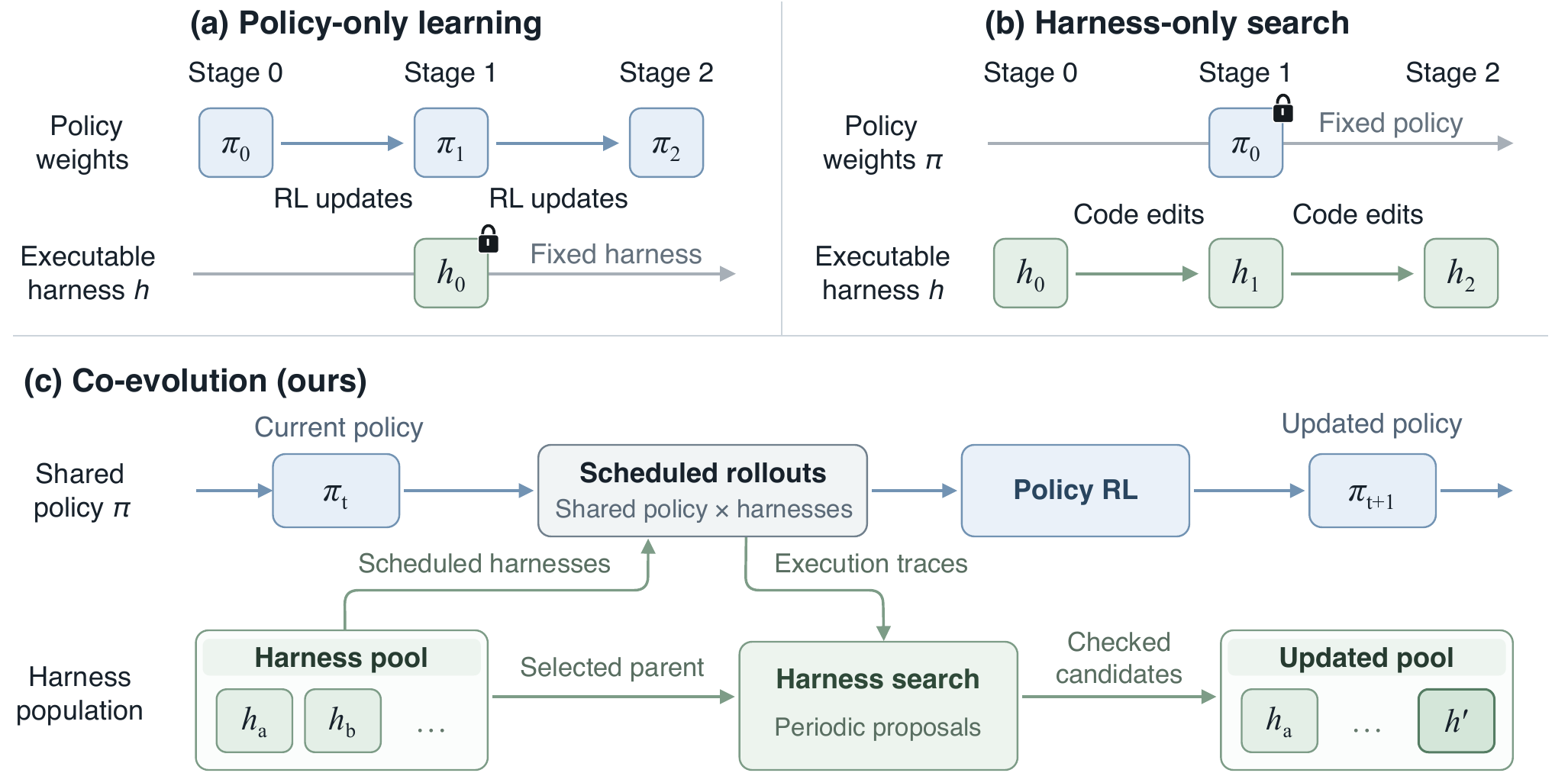}
\caption{Coupling policy optimization with harness evolution.
Policy-only RL adapts policy weights under a fixed harness (a), whereas
harness-only search adapts the harness under a fixed policy (b).
\method{} jointly adapts a shared policy and a harness population (c),
using execution traces to guide harness edits and population rollouts
to update the policy.}
\label{fig:motivation}
\end{figure}

The two forms of adaptation are also linked during training.
Repairing a tool handler changes the feedback available to the policy
and the trajectories from which it learns. As the policy improves,
its execution traces can reveal further harness weaknesses that were
not apparent under the initial model. Searching once before RL cannot
respond to those later failures; keeping the policy frozen cannot
train it on the experience collected through the revised interface.
We therefore ask (Figure~\ref{fig:motivation}): how can an agent's policy
and execution interface adapt together throughout training?

Recent methods update both model weights and harness or training
code~\citep{evotrainer,hase}. WHALE alternates harness search under
frozen model weights with rejection-sampling fine-tuning under a
selected harness~\citep{whale}. Its policy updates therefore use one
selected execution interface at a time. In \method{}, multiple
harnesses remain active during training: a scheduler allocates
rollouts across the evolving population, and these rollouts jointly
update one shared policy through reinforcement learning.
This allows new candidates to contribute training experience alongside
existing versions, rather than replacing the interface for all
subsequent rollouts. We evaluate the benefit of this population-based
design against single-harness alternating updates under the same
training budget.

\begin{figure}[t]
\centering
\vspace{-1em}
\includegraphics[width=\linewidth]{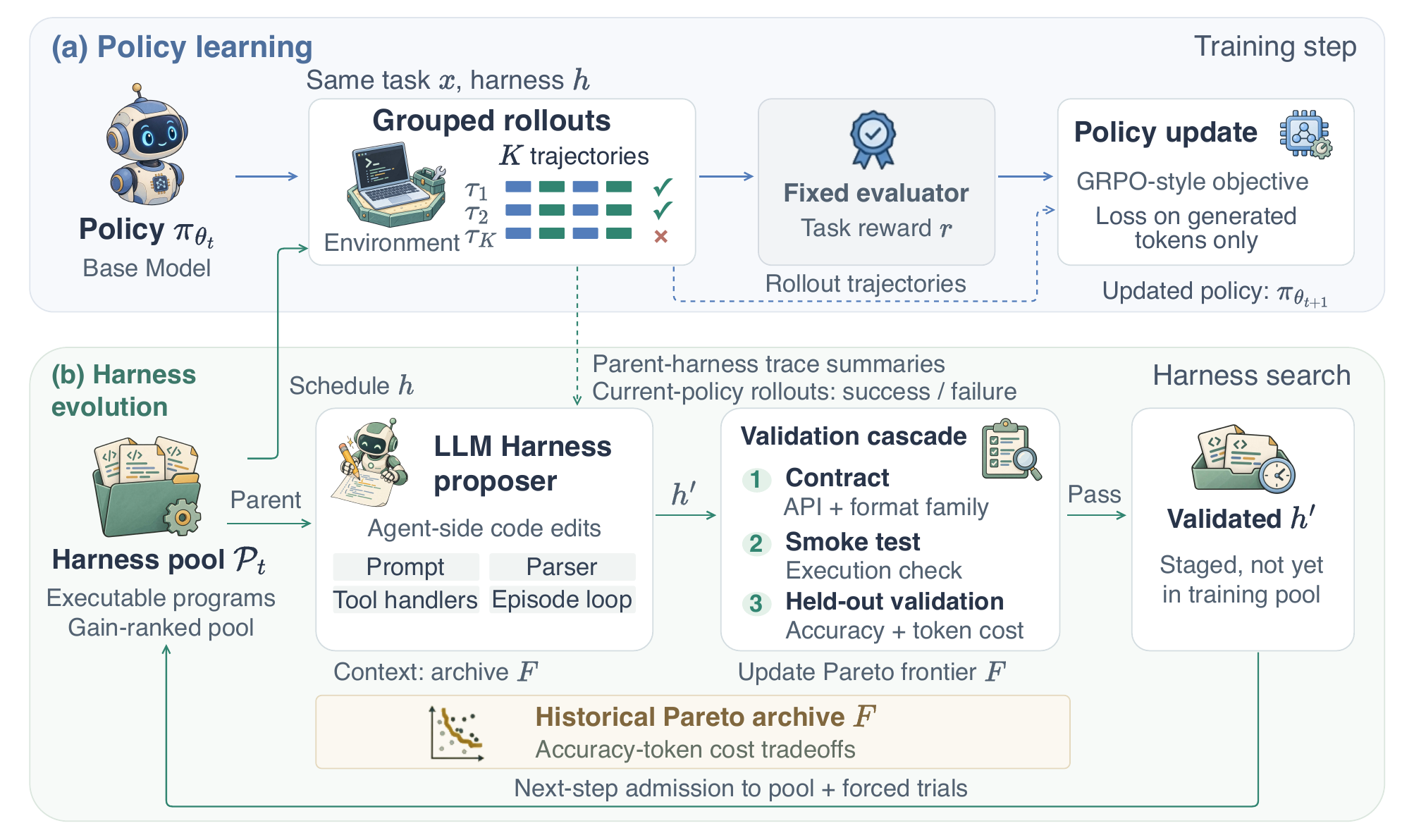}
\caption{Overview of \method{}. A shared policy learns from population
rollouts, while a proposer revises harness code using current-policy
traces and historical accuracy--cost trade-offs. Accepted candidates
enter the pool at step boundaries. The validation cascade shown applies
to TB-2; SWE uses contract checks followed by pool trials. Blue and green
rollout blocks represent generated tokens and tool observations,
respectively; only generated tokens contribute to the policy loss.}
\label{fig:overview}
\vspace{-1em}
\end{figure}

In \method{}, a proposer model reads the policy's successful and failed
trajectories and revises harness code. Accepted candidates join the
population, from which a scheduler selects versions for training
rollouts. The proposer also receives a Pareto archive of historical
accuracy--cost trade-offs. Figure~\ref{fig:overview} shows how search
and rollout collection alternate with policy updates.

We evaluate \method{} on software-engineering and terminal-based tasks.
On SWE-rebench, it achieves a 27.6\% mean
resolved rate, compared with 21.2\% for harness-only search and 20.3\%
for policy-only training. \method{} also outperforms both
baselines on SWE-bench Verified without further adaptation. A separately
trained Terminal-Bench~2.0 agent improves on harness-only search.
Matched-budget controls favor continued population-based training over
one-time search and single-harness alternation; ablations examine
execution feedback and historical context.

Our contributions are threefold: (i) a population-based formulation of
policy--harness co-evolution, in which diverse execution interfaces supply
experience to one shared policy; (ii) a search-and-training procedure
that uses current-policy traces and historical accuracy--cost context
to propose programs, then schedules accepted versions for future rollouts;
and (iii) evaluation across three benchmarks, with matched-budget
controls for one-time search, single-harness alternation, and search context.

\section{Related Work}
\label{sec:related}
\paragraph{Automated Agent and Harness Design.}
Agent design methods optimize executable code~\citep{adas,dgm}, prompts~\citep{gepa},
and reusable context or skills~\citep{ace,evoharness,skilloptlite}.
Execution-feedback search~\citep{metaharness,selfharness,autosaddler,harnessevolve},
structured edit spaces~\citep{harnesscompass,modularrsi,xia2026rrsi}, and learned
editors~\citep{harnessr1} are used to search for better designs.
Evaluation studies examine generalization, cost, and differences across
policies~\citep{harnessdev,rethinkingharness}. We evolve
harnesses during RL, when their outputs also become training data.

\paragraph{Reinforcement Learning for LLM Agents.}
Reward-based training improves coding policies~\citep{swerl,deepswe},
with complementary work on credit assignment~\citep{harnessrl_blackbox}
and RL through deployed or black-box harnesses~\citep{agentlightning,clawgym2}.
Other methods adapt skills~\citep{skillrl,metaclaw} or runtime
state management~\citep{evoharnessrl}. We use group-relative
optimization~\citep{deepseekmath} to train one policy on rollouts from
multiple evolving harnesses; we do not change the gradient estimator.

\paragraph{Policy--Harness Co-Evolution.}
Joint adaptation includes harness revision~\citep{harnessx,coharness},
expert correction~\citep{onpolicycorrection}, and verified data
generation~\citep{helix}. Designs differ in adaptation targets and
policy sharing: EvoTrainer evolves training-side
infrastructure~\citep{evotrainer}; WHALE alternates model updates and
harness search~\citep{whale}; HASE uses one model for solving and
editing~\citep{hase}; and HarnessForge maintains harness-specific
adapters~\citep{harnessforge}. \method{} instead uses a separate proposer
to evolve agent-side programs while training one shared policy across
the scheduled population. Unlike HarnessBandit's scheduling over a
given harness set~\citep{harnessbandit}, our search also changes the
programs available during training. Appendix~\ref{sec:comparison}
details these distinctions.

\section{Method}
\label{sec:method}
\suppressfloats[t]

\method{} alternates rollout collection and policy optimization with
periodic harness search (Figure~\ref{fig:overview}). One policy is shared
across all active harnesses. The scheduler selects rollout harnesses
from the training pool, and the proposer edits their code. The proposer
also reads historical accuracy--cost records from a separate archive,
which can retain information about versions no longer in use.

\subsection{Agents as Coupled Weight--Harness Systems}
\label{sec:method:setup}

Let $\pi_\theta$ be the policy and $h$ an executable harness. On task $x$,
an episode $\tau\sim\mathrm{Rollout}(\pi_\theta,h,x)$ receives task reward
$r(\tau)\in[0,1]$ from a fixed evaluator. The deployment objective is
\begin{equation}
 \max_{\theta,h}J(\theta,h),\qquad
 J(\theta,h)=\mathbb E_{x\sim\mathcal D}
 \mathbb E_{\tau\sim\mathrm{Rollout}(\pi_\theta,h,x)}[r(\tau)].
 \label{eq:joint}
\end{equation}
\method{} maintains a harness training pool $\mathcal P_t$ and a separate archive
$F_t$ of non-dominated recorded evaluations. Let
$q_t(h\mid x,\mathcal P_t)$ describe the scheduler's rollout assignment.
The corresponding task-reward objective under this assignment is
\begin{equation}
 \bar J_t(\theta)=\mathbb E_{x\sim\mathcal D_{\rm train}}
 \sum_{h\in\mathcal P_t}q_t(h\mid x,\mathcal P_t)
 \mathbb E_{\tau\sim\mathrm{Rollout}(\pi_\theta,h,x)}[r(\tau)].
 \label{eq:popavg}
\end{equation}
The scheduler defines the mixture of execution interfaces seen during
training; this mixture is uniform only under uniform assignment.
Assignment and harness code remain fixed within each update.
Equation~\eqref{eq:popavg} describes the scheduled training distribution,
whereas Eq.~\eqref{eq:joint} evaluates a deployed policy--harness pair.
Both use task reward $r$ to characterize performance; the implemented
policy update uses the domain-specific training reward $R$, including
SWE reward shaping (Appendix~\ref{sec:reward-protocol}). Neither the
scheduled mixture nor the shaped surrogate is assumed to share the
optimum of the deployment objective.

\subsection{The \method{} Training Loop}
\label{sec:method:loop}

Algorithm~\ref{alg:coevolve} gives the training loop. Each training step
collects task--harness rollout groups and updates the shared policy if
enough usable samples remain. Search uses the collected trajectories;
accepted candidates become available at the next step boundary.
A new version is never applied to an already collected batch. A training
step can finish without an optimizer update if batch preparation leaves
too few usable samples.

\begin{algorithm}[t]
\caption{\method{}: Co-evolution of policy weights and a harness population}
\label{alg:coevolve}
\begin{algorithmic}[1]
\Require Initial policy $\pi_{\theta_0}$, seed harness $h_0$,
  tasks $\mathcal{D}$, group size $K$, batch size $B$, harness slots per step
  $N$, pool $\mathcal{P} \gets \{h_0\}$, evaluated archive $F$, $\mathrm{Staged}\gets\varnothing$
\For{$t = 0, 1, \dots, T-1$}
  \State $\mathcal{P} \gets \mathrm{Admit}(\mathcal{P},\mathrm{Staged})$; $\mathrm{Staged}\gets\varnothing$
  \State $(h_1,\dots,h_N) \gets \mathrm{Schedule}(\mathcal{P})$
  \State sample $\{x_b\}_{b=1}^{B} \sim \mathcal{D}$; assign each task a
    harness $h(b)$
  \For{$b = 1, \dots, B$ \textbf{in parallel}}
    \State $\tau_{b,1},\dots,\tau_{b,K} \sim
      \mathrm{Rollout}(\pi_{\theta_t}, h(b), x_b)$
  \EndFor
  \State $\mathcal B_t\gets\mathrm{PrepareBatch}(\{\tau_{b,i}\}_{b,i})$
  \State $\theta_{t+1}\gets\theta_t$
  \If{$\mathcal B_t$ contains enough usable samples}
    \State compute group-relative advantages using Eq.~\eqref{eq:adv}
    \State $\theta_{t+1} \gets \mathrm{PolicyUpdate}(\theta_t,\mathcal B_t)$
  \EndIf
  \State update $\{g_h\}$ and $F$ from available evaluations; evict eligible members
  \If{domain-, phase-, and pool-dependent proposal cadence permits}
    \State $h_{\mathrm{par}} \gets \mathrm{SelectParent}(\mathcal{P})$
    \State $h' \sim q_{\mathrm{LLM}}\!\left(\cdot \mid h_{\mathrm{par}},
      \mathcal{T}_{\theta_t}(h_{\mathrm{par}}), F\right)$
    \If{$h'$ passes the domain-specific candidate checks}
      \State $\mathrm{Staged} \gets \mathrm{Staged} \cup \{h'\}$
    \EndIf
  \EndIf
  \State every $10$ training steps: monitor on validation tasks
\EndFor
\end{algorithmic}
\end{algorithm}

\paragraph{Policy learning across the harness population.}
Each sampled group targets $K$ rollouts for the same task and harness.
Rewards are centered within this fixed task--interface pair, rather than
across trajectories collected with different harnesses.
After batch preparation, let $\mathcal I_b$ index the retained episodes
in group $b$, with $k_b=|\mathcal I_b|$. Let $R_{b,i}$ denote the
training reward after the applicable reward transformations, rather than
the separate filtering score (Appendix~\ref{sec:reward-protocol}).
The group-relative advantage is
\begin{equation}
 \hat A_{b,i}=R_{b,i}-\frac1{k_b}\sum_{j\in\mathcal I_b} R_{b,j},
 \qquad i\in\mathcal I_b.
 \label{eq:adv}
\end{equation}
Appendix~\ref{sec:protocols} specifies batch preparation.
For the clipped update used in the SWE configuration, let
$u_{b,i,\ell}(\theta)=\pi_\theta(a_{b,i,\ell}\mid H_{b,i,\ell})/
\pi_{\theta_{\rm old}}(a_{b,i,\ell}\mid H_{b,i,\ell})$, where
$H_{b,i,\ell}$ precedes generated token $a_{b,i,\ell}$ and
$\theta_{\rm old}$ denotes the rollout policy parameters.
Let $\mathcal L_{b,i}$ index generated tokens in trajectory $\tau_{b,i}$
and $L=\sum_{b,i}|\mathcal L_{b,i}|$ over the trajectories included in
the update. The surrogate is
\begin{equation}
 \mathcal J_{\rm GRPO}(\theta)=\frac1L\sum_b\sum_{i\in\mathcal I_b}\sum_{\ell\in\mathcal L_{b,i}}
 \min\!\left\{u_{b,i,\ell}\hat A_{b,i},
 \operatorname{clip}(u_{b,i,\ell},1-\epsilon_-,1+\epsilon_+)\hat A_{b,i}\right\}.
 \label{eq:grpo}
\end{equation}
Only generated tokens contribute; tool observations are conditioning
context. The clipping parameters $\epsilon_-$ and $\epsilon_+$ are
specified in Table~\ref{tab:hparams}. This token-normalized surrogate uses
the scheduled harness mixture, not a uniform average over pool members.

\paragraph{Harness search from execution traces.}
Harness versions are self-contained programs exposing a fixed episode
API. A more capable proposer model acts as a stochastic mutation operator,
$h' \sim q_{\mathrm{LLM}}(\cdot \mid h_{\mathrm{par}},
\mathcal{T}_{\theta_t}(h_{\mathrm{par}}), F)$, and may rewrite files in
the parent version $h_{\mathrm{par}}$
(prompt, parser, tool handlers, or the rollout loop itself),
conditioned on the parent's source, summaries
$\mathcal{T}_{\theta_t}(h_{\mathrm{par}})$ of the \emph{current} policy's
successful and failed traces under the parent, the Pareto frontier $F$,
and same-round rejected attempts. Two hard constraints apply: the
proposer cannot modify the training framework, and it cannot change the
action-format family used by the policy. Candidate checks are
domain-specific: SWE uses a structural contract check followed by pool
trials, while TB-2 additionally uses smoke testing and candidate
validation before pool trials. Accepted versions are staged and admitted
only at step boundaries. Trial rollouts provide initial usage statistics
and update their ranking. Appendix~\ref{sec:protocols} details these paths.

Search and policy optimization operate on different objects: the proposer
edits execution code from trajectory feedback, while RL updates the shared
weights from population rollouts. An accepted edit adds an alternative
interface for subsequent collection; it does not replace every active
harness. Its usefulness is then assessed through pool trials under the
training policy.

\paragraph{Pool scheduling.}
Pool members may encounter tasks of different difficulty, making raw
reward averages unsuitable for direct comparison. The scheduler ranks members by a
difficulty-adjusted performance score with a lower-confidence-bound guard:
\begin{equation}
  g_h \;=\; \mathrm{EWMA}\Big[\, \overline{\big(r - b_x\big)\, w_x}
  \,\Big],
  \qquad
  g_h^{\mathrm{lcb}} \;=\; g_h - \frac{c}{\sqrt{\max(n_h,1)}},
  \label{eq:gain}
\end{equation}
where $b_x$ is a per-task success-rate baseline (an EWMA over that task's
history), the overbar averages observations for harness $h$,
$w_x$ is a task-difficulty weight, $n_h$ the member's use count,
and $c$ a conservative ranking penalty. The LCB term lowers the score
of under-sampled members; protection from premature eviction instead
comes from forced trials and a minimum-use rule. This is a heuristic
score rather than a calibrated statistical confidence interval.
It governs harness selection, eviction, and parent choice, but is not
added to the training advantage in Eq.~\eqref{eq:adv}. The scheduler
thus changes which experience is collected without redefining the
reward used by the policy update.
Scheduling combines performance-based allocation with candidate trials.
The scheduler allocates $N$ slots, not necessarily $N$ distinct versions,
so training can start from the seed alone. Newly admitted candidates
receive forced trials and remain protected until the minimum-use
requirement is met. Their rollout outcomes update the ranking used for
later allocation and eviction. Parent selection draws from the active
pool using these performance scores; retaining a version in the archive
alone does not make it eligible as a parent.

\paragraph{Historical accuracy--cost archive.}
The archive retains non-dominated recorded accuracy--cost pairs:
a version dominates another when its accuracy is no lower and its
mean generated-token cost no higher, with at least one strict inequality.
The pool and archive serve different decisions. The pool determines
which programs collect training experience; the archive supplies the
proposer with previously observed accuracy--cost trade-offs, including
records of versions evicted from the pool. Parent traces describe how
the current policy uses one program, whereas archive records provide
context across earlier evaluations. These records retain their historical
scores rather than estimating every version's performance under the
current policy. Appendix~\ref{sec:proofs} gives the fixed-score properties.

\subsection{Analysis of Task-Reward Contrasts}
\label{sec:method:why}
Identical rewards within a task--harness group give zero advantages in
Eq.~\eqref{eq:adv}. For $K\ge2$ conditionally independent binary outcomes
with success probability $p$, the probability of a non-uniform group is
\begin{equation}
\rho_K(p)=1-p^K-(1-p)^K.
\label{eq:contrast}
\end{equation}
For rare success, $\rho_K(p)\approx Kp$: improving the interface can
make reward contrasts more frequent on difficult tasks. The relation
reverses above $p=1/2$, so higher success does not always yield more
contrast. This provides an intuition for coupling harness adaptation
with group-relative learning, not a convergence guarantee.
Appendix~\ref{sec:proofs} gives the derivation; SWE's shaped filtering
scores are treated separately in Appendix~\ref{sec:retention}.

\subsection{Task Rewards and Evaluator Integrity}
\label{sec:method:reward}

Joint adaptation creates opportunities for reward exploitation through
both policy behavior and harness code. For SWE evaluation, success is
strictly binary, $r{=}1$ iff all
\textsc{fail\_to\_pass} tests pass and all \textsc{pass\_to\_pass} tests
continue to pass, with forced submission at episode end and no LLM judge.
SWE training combines test correctness with submission-gated process
rewards and applies a soft length penalty before the policy update.
Group filtering uses the separate \texttt{test\_score} field, whose
meaning depends on the harness: it contains only test pass fractions
in the seed harness, but can include process rewards in evolved versions.
Thus, evaluation uses strict resolution, policy advantages use the
transformed training reward, and retention diagnostics use the filtering
score. A retained SWE group need not contain a fully resolved episode.
Appendix~\ref{sec:reward-protocol} gives the reward and score mapping.
TB-2 instead uses binary training rewards, without soft length penalties
or per-harness reward normalization.

Harness edits could expose hidden tests to the policy. The seed harness
disables hidden-test injection, and an audit gate rejects detected
attempts to reintroduce access. Code inspection found no hidden-test
access in five admitted harnesses; a separate probe rejected six
test-access proposals. Appendix~\ref{sec:leakage} describes the audit
scope and limitations.

\section{Experiments}
\label{sec:experiments}

Our experiments address three questions:
1) Does co-evolution improve task performance over harness-only search
and policy-only RL, and do the gains extend beyond SWE-rebench?
2) What do online search, the harness population, and proposer context
contribute under matched training budgets?
3) How do training rewards and harness utilization evolve during learning?

\subsection{Setup}
\label{sec:exp:setup}

\paragraph{Benchmarks and metrics.}
Our primary experiments use SWE-rebench~\citep{swerebench}, with tasks
dated before 2026 for training and 110 tasks dated after March 2026 for
held-out evaluation. We report mean resolved rates over three attempts
per task (avg@3; temperature~0.7). To assess cross-dataset generalization,
we evaluate the SWE-rebench-trained agents on the 500-task SWE-bench
Verified benchmark~\citep{swebench,verified}.
We also train a separate agent on Terminal-Bench~2.0
(TB-2)~\citep{terminalbench}, using Harbor~\citep{Harbor_Framework} for evaluation and Terminus-2
as the baseline harness. SWE tasks are scored by strict resolution;
TB-2 tasks use their benchmark-provided evaluators.

\paragraph{Model and training.}
We train Qwen3.5-35B-A3B~\citep{qwen35blog} directly with RL using
rank-32 LoRA~\citep{lora} based on SkyRL~\citep{griggs2025skyrl}.
Harness scheduling and code search are integrated into rollout
collection. For SWE, we use group-relative advantages with asymmetric
importance-ratio clipping~\citep{dapo}, without a reference-policy
reward penalty or separate KL loss.
SWE-rebench training runs for 150 steps with a target of 48 rollouts
per step, distributed across a pool of up to eight harness versions.
At each search invocation, Claude Opus 4.7~\citep{claudeopus47} proposes
up to two candidates, which are checked as described in
\S\ref{sec:method:loop}. \method{} uses approximately as many training
episodes as the RL-only baseline, with additional computation for
harness validation and proposer inference. Full training and evaluation
settings, including search schedules and costs, are provided in
Appendix~\ref{sec:protocols}.

\paragraph{Baselines.}
We compare \method{} with harness-only search and policy-only RL.
For harness-only search, we reproduce Meta-Harness~\citep{metaharness}
starting from SWE-mini, with the initial policy held fixed.
For policy-only RL, Plain GRPO trains the policy with SWE-mini held fixed.
The initial policy paired with SWE-mini serves as the Base baseline.
Our ablation study further compares two ways to combine search and RL:
Search-then-RL selects a harness before policy training and keeps it
fixed thereafter; single-harness alternating RL interleaves policy
training with revisions to a single active harness.

\begin{figure}[t]
\centering
\vspace{-1em}
\includegraphics[width=.85\linewidth]{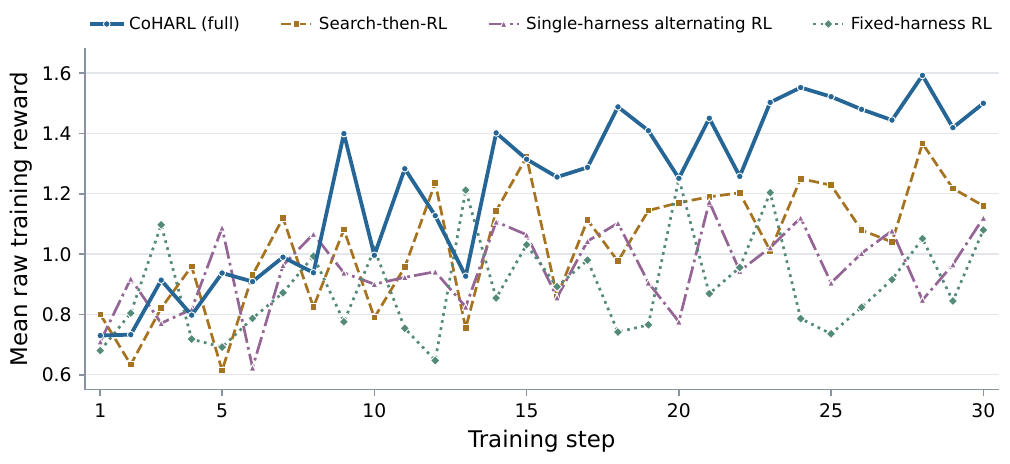}
\caption{Training rewards for the full method and three controls in
Table~\ref{tab:ablation}. Each point is the mean raw reward over retained
training samples, before length penalties, KL adjustments, or normalization.
Curves show unsmoothed values, not evaluation resolved rates.}
\vspace{-1em}
\label{fig:training-curves}
\end{figure}

\subsection{Main Results}
\label{sec:exp:main}
\paragraph{SWE-rebench performance.}
\method{} achieves a 27.6\% resolved rate, outperforming Meta-Harness
by 6.4 percentage points and Plain GRPO by 7.3 points
(Table~\ref{tab:main}). Both baselines improve on the initial agent,
with Meta-Harness achieving a higher resolved rate than Plain GRPO
without changing the policy weights. Improving the execution interface
can therefore be effective even without policy training, but neither
interface optimization nor policy-only RL matches joint adaptation.
In \method{}, harness revisions also change the experience available
for subsequent policy updates. Its advantage over
both baselines is consistent with treating the harness as part of the
learning process rather than only an inference-time wrapper.

\begin{table}[!ht]
  \centering
  \caption{Performance on the 110-task SWE-rebench evaluation split.
  Resolved rates are averaged over three attempts per task (avg@3)
  at temperature 0.7. $^{*}$Our Meta-Harness reproduction, initialized from SWE-mini.}
  \label{tab:main}
  \setlength{\tabcolsep}{5pt}
  \begin{tabular}{@{}lllr@{}}
    \toprule

    \textbf{Method} & \textbf{Weights} &
    \textbf{Harness} & \textbf{Resolved} \\
    \midrule
    Base         & Qwen3.5-35B-A3B  & SWE-mini     & 17.3\% \\

    Meta-Harness & Qwen3.5-35B-A3B  & Meta-Harness$^{*}$ & 21.2\% \\
    Plain GRPO   & RL-trained & SWE-mini     & 20.3\% \\

    \method{} (ours) & RL-trained     & Co-evolved   & \best{27.6\%} \\
    \bottomrule
  \end{tabular}
\end{table}

\paragraph{Cross-dataset generalization.}
\label{sec:exp:transfer}

On SWE-bench Verified, the policy and harness trained on SWE-rebench
resolve 59.8\% of tasks without further adaptation
(Table~\ref{tab:verified}). All four methods retain their relative
ordering, and \method{}'s lead over both optimization baselines is larger
in absolute percentage points than on SWE-rebench. The learned
policy--harness combination retains its advantage on another task
collection without benchmark-specific adaptation.
For comparison with closed-source agents, published results include
33.2\% for GPT-4o~\citep{openai2025gpt41} and 49.0\% for
Claude 3.5 Sonnet~\citep{anthropic2025swe}.
\method{} reaches 59.8\% with an open-weight policy, comparing favorably
with these earlier systems. These published results provide an external
reference rather than a controlled model comparison: harnesses and
inference settings differ, and the OpenAI evaluations exclude 23 of
the 500 tasks.

\begin{table}[!ht]
  \centering
  \caption{Resolved rates on the 500-task SWE-bench Verified benchmark
  with one attempt per task. SWE-rebench-trained agents are evaluated
  without further adaptation.
  $^{\dagger}$Provider-reported results under original evaluation
  settings~\citep{openai2025gpt41,anthropic2025swe}.
  $^{*}$Our Meta-Harness reproduction, initialized from SWE-mini.}
  \label{tab:verified}
  \setlength{\tabcolsep}{5pt}
  \begin{tabular}{@{}lllr@{}}
    \toprule
    \textbf{Methods} & \textbf{Weights} & \textbf{Harness} & \textbf{Resolved} \\
    \midrule
    \footnotesize \textit{Closed-source models} & & & \\
    \addlinespace[2pt]
    GPT-4o$^{\dagger}$ & - & - & 33.2\% \\
    Claude 3.5 Sonnet$^{\dagger}$ & - & - & 49.0\% \\
    \midrule
    \multicolumn{4}{@{}l}{\footnotesize \textit{Qwen3.5-35B-A3B-based methods}} \\
    \addlinespace[2pt]
    Base & Fixed & SWE-mini & 44.2\% \\
    Meta-Harness & Fixed & Meta-Harness$^{*}$ & 52.0\% \\
    Plain GRPO & RL-trained & SWE-mini & 49.8\% \\
    \method{} (ours) & RL-trained & Co-evolved & \best{59.8\%} \\
    \bottomrule
  \end{tabular}
\end{table}

\paragraph{Application to terminal agents.}
\label{sec:exp:tb2}

On Terminal-Bench~2.0, a separately trained \method{} agent achieves a
36.8\% success rate, exceeding harness-only search by 3.1 percentage
points (Table~\ref{tab:tb2}). Harness-only search already improves on
Base, showing that revising the execution interface benefits terminal
task solving. Co-evolution improves further by allowing policy learning
alongside harness search, rather than relying on interface changes alone.

\begin{table}[!ht]
  \centering
  \caption{Task success rates on Terminal-Bench~2.0, evaluated through
  Harbor with Terminus-2 as the baseline harness. For our evaluations at
  temperature 0.6, success is averaged within each task over returned
  episodes and then across tasks; tasks with no returned episode score zero.
  $\dagger$: Shepherd~\citep{shepherd}, 89 tasks, avg@5.}
  \label{tab:tb2}
  \setlength{\tabcolsep}{5pt}
  \begin{tabular}{@{}llr@{}}
    \toprule

    \textbf{Method} & \textbf{Harness} &
    \textbf{Success} \\
    \midrule
    Base                  & Terminus-2   & 26.6\% \\

    Meta-Harness          & Meta-Harness & 33.7\% \\
    Flat GRPO (Shepherd)$^{\dagger}$ & Terminus-2   & 34.2\% \\

    \method{} (ours)         & Co-evolved   & \best{36.8\%} \\
    \bottomrule
  \end{tabular}
\end{table}
\subsection{Analysis}
\label{sec:exp:analysis}
\label{sec:exp:ablation}

\paragraph{Ablation study.}
We ablate harness evolution and proposer context, and compare alternative
training strategies under matched 30-step training budgets
(Table~\ref{tab:ablation}). Fixing the seed harness (A1) reduces the
resolved rate from 25.5\% to 20.0\%, showing the benefit of adapting
the interface during policy training. Removing archive context (A2) or
execution traces (A3) also lowers performance, even though both variants
retain online search. These inputs serve different purposes: the archive
records previously observed accuracy--cost trade-offs, whereas traces
show how the current policy uses the parent harness. The two ablations
indicate that both sources of context help the proposer produce useful
revisions.

\begin{table}[!ht]
  \centering
  \caption{Component ablations and training-strategy comparisons on
SWE-rebench under matched training budgets (30 steps).
We report strict resolved rates and counts over 110 tasks with one attempt per task.}

  \label{tab:ablation}
  \setlength{\tabcolsep}{5pt}
  \begin{tabular}{@{}lrr@{}}
    \toprule

    \textbf{Configuration} & \textbf{Resolved rate} &
    \textbf{Resolved} \\
    \midrule
    \method{} (full)         & \best{25.5\%} & 28/110 \\
    A1: Fixed-harness RL & 20.0\% & 22/110 \\
    A2: No archive context  & 20.0\% & 22/110 \\
    A3: No execution traces & 20.9\% & 23/110 \\
    A4: Search-then-RL          & 22.7\% & 25/110 \\
    A5: Single-harness alternating RL & 19.1\% & 21/110 \\
    \bottomrule
  \end{tabular}
\end{table}

Search-then-RL (A4) improves on fixed-harness RL but remains 2.7
percentage points below the full method. Its harness is selected for
the initial policy and remains fixed throughout RL, even as the policy's
capabilities and execution behavior change. Co-evolution instead allows
harness search to respond to the updated policy, potentially finding
interfaces that better exploit its improving capabilities.
Single-harness alternating RL (A5) updates both
components online yet trails the full method by 6.4 points. Unlike the
population-based method, each accepted revision replaces the sole
interface used for subsequent rollouts. Such replacements may abruptly
shift both the training data distribution and the reward distribution,
potentially disrupting policy optimization.
Retaining multiple harnesses may moderate these shifts by continuing to
collect experience through previously used interfaces. Search schedules
and candidate acceptance rules are detailed in Appendix~\ref{sec:protocols}.

\paragraph{Training dynamics.}

The training reward curves in Figure~\ref{fig:training-curves} provide
a temporal view of the same 30-step comparisons. The methods overlap
early, while \method{} develops a lead later and remains above all three
controls during the final ten steps despite step-to-step fluctuations.
This late-training
reward advantage accompanies the higher final resolved rate in
Table~\ref{tab:ablation}, linking the endpoint comparison to differences
that emerge over the course of training.

\begin{figure}[!htbp]
\centering
\includegraphics[width=\linewidth]{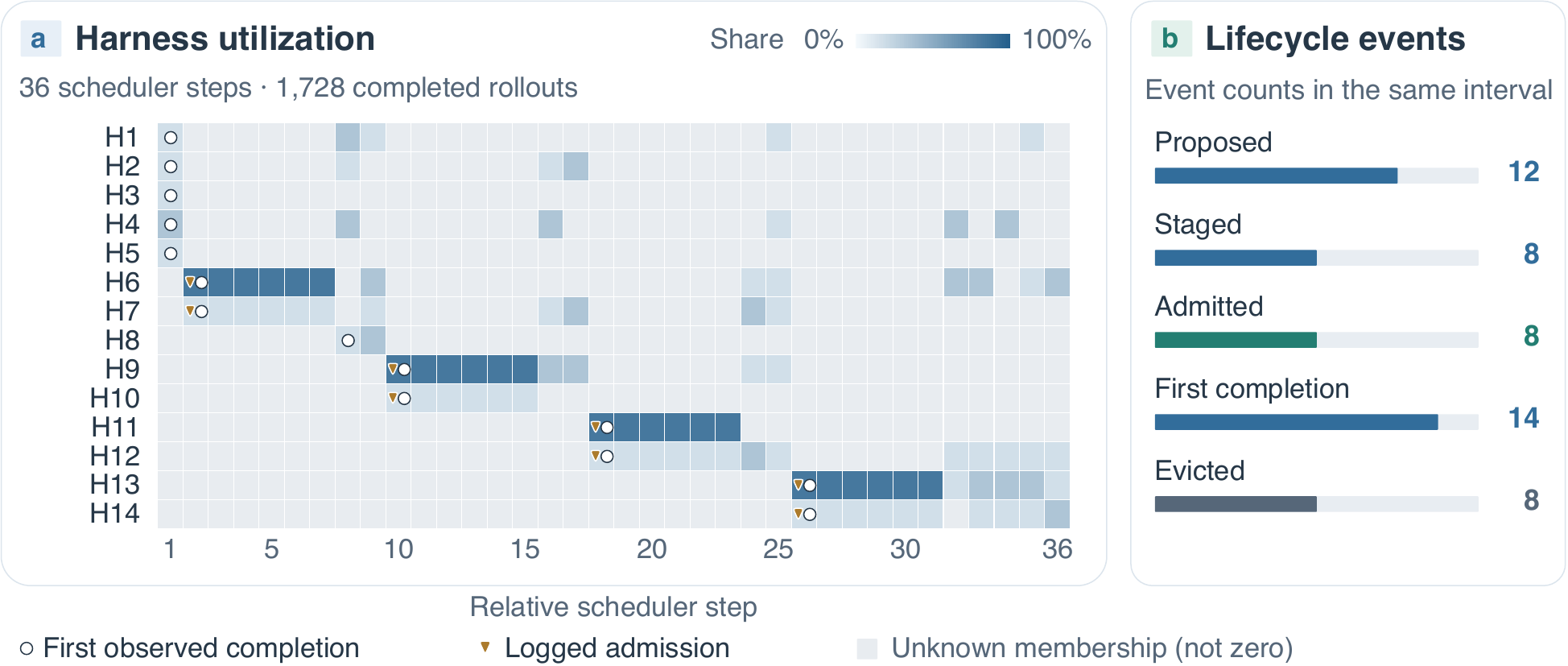}
\caption{Harness utilization and population turnover.
(a) Per-step rollout shares across harness versions; darker blue indicates
larger shares, while gray denotes unknown pool membership.
(b) Recorded lifecycle events over the same 36-step segment.}
\label{fig:dynamics}
\end{figure}

\paragraph{Harness utilization.}
To examine how the population is used, Figure~\ref{fig:dynamics} tracks
14 harness versions over 36 scheduler steps from a separate archived run.
Several versions contribute rollouts within the same step, and earlier
harnesses remain in use after newer ones enter the pool. The population
is therefore used for concurrent experience collection, rather than
merely storing past revisions. The contribution of each harness to the
collected training experience changes over time.
In contrast to single-harness alternating RL,
admitting a candidate does not require replacing the interface for all
subsequent rollouts. The lifecycle counts complement this view by
summarizing admissions and evictions; Appendix~\ref{sec:archival} gives
the counting rules.

\section{Conclusion}
\label{sec:conclusion}

We presented \method{}, which jointly trains a shared policy and evolves
a population of executable harnesses. Experiments on SWE-rebench,
SWE-bench Verified, and Terminal-Bench~2.0 demonstrate the benefits of
adapting both components. Matched-budget ablations favor continued
population-based evolution over searching once before RL or repeatedly
revising a single harness. Together, these results show that executable
interfaces are an important target for agent learning alongside policy
weights: policy behavior informs harness revisions, and revised harnesses
shape subsequent training experience.

\subsection*{AI use statement}
We used GPT-6-Astra to edit and organize the manuscript, revise
mathematical statements and proofs, interpret results, obtain feedback
on experimental design, and check literature and references.
We also used AI assistance for figure layouts, visual design drafts,
and icon generation.
The authors reviewed the AI-assisted revisions and checked the
mathematical arguments, experimental interpretations, and references.
The authors take responsibility for the final text, claims, and artifacts.

\subsection*{Ethics statement}
Executable agents can expose repositories and environments to unintended
changes. The method uses isolated execution and restricts proposer edits,
but finite audits do not prove security. Any release should respect
benchmark and repository licenses, remove credentials and identifying
information from traces, and document risks of executing evolved programs.

\subsection*{Reproducibility statement}
Algorithm~\ref{alg:coevolve} describes the procedure,
Appendix~\ref{sec:proofs} gives assumptions and proofs, and
Appendix~\ref{sec:appendix} records the reported hyperparameters.
Appendix~\ref{sec:protocols} specifies reward computation, batch preparation,
candidate checks, and evaluation protocols. A complete versioned artifact
bundle is not included with this manuscript.

\bibliographystyle{iclr2027_conference}

\bibliography{references}

\newpage

\appendix

\section{Proofs}
\label{sec:proofs}

For a fixed task, policy, and harness, let $p$ be the probability of
binary task success.
\begin{proposition}[Non-uniform binary-reward groups]
\label{prop:flywheel}
For $K\ge2$ conditionally independent Bernoulli$(p)$ rollouts,
$\rho_K(p)=1-p^K-(1-p)^K$ is the probability of non-uniform raw
task rewards. For fixed $K$ and $p\ll1/K$, $\rho_K(p)\approx Kp$.
If an edit increases $p$ to $cp$ with $c>1$ and $cp\ll1/K$, the
frequency of such groups increases by approximately $c$.
\end{proposition}
The proposition identifies a regime in which better harnesses can make
task-reward contrasts more frequent. This relationship is not globally
monotone: the probability increases with $p$ only below $1/2$ and has
maximum $1-2^{1-K}$. It characterizes binary task-reward variation,
rather than policy-gradient magnitude or the nonbinary SWE
filtering-score diagnostics.
For heterogeneous tasks, the relevant expectation is
$\mathbb E_x[\rho_K(p_x)]$, not $\rho_K(\mathbb E_x[p_x])$.
Thus, improvements on low-success-probability tasks can increase the
frequency of groups with a usable task-reward contrast.

\begin{proof}[Proof of Proposition~\ref{prop:flywheel}]
Condition on a fixed task, policy, and harness, and assume $K\ge2$
independent Bernoulli$(p)$ outcomes. All-failure and all-success groups
have probabilities $(1-p)^K$ and $p^K$. These events are disjoint;
their complement has probability $1-p^K-(1-p)^K$. Only on that
complement are the mean-centered raw rewards nonzero.
For fixed $K$, expansion gives $\rho_K(p)=Kp+O(p^2)$ as $p\to0$.
Also $\rho'_K(p)=K((1-p)^{K-1}-p^{K-1})$ is positive below $1/2$
and negative above it, so $\max_p\rho_K(p)=1-2^{1-K}$.
This concerns raw reward variation, not the magnitude of the policy gradient.
\end{proof}

\paragraph{Historical accuracy--cost archive.}
Write $a(h)$ and $c(h)$ for recorded evaluation accuracy and mean
generated-token cost. A version dominates another if its accuracy is
no lower and cost no higher, with at least one strict inequality.
Let $V_n$ contain all versions with recorded evaluation scores and
$F_n=\mathrm{ND}(V_n)$ be its non-dominated subset.
\begin{proposition}[Properties of a fixed-score archive]
\label{prop:pareto}
Suppose scores are fixed, $V_n$ is finite, and $V_n\subseteq V_{n+1}$.
(i) No member of $F_n$ strictly dominates another; identical score
pairs may coexist. (ii) A candidate dominated by a recorded parent
cannot belong to $F_n$ while that parent remains in $V_n$.
(iii) For any budget $C$, the best recorded accuracy within budget,
$\max\{a(h):h\in F_n,c(h)\le C\}$, is non-decreasing in $n$,
taking the empty-set maximum as $-\infty$.
\end{proposition}

\begin{proof}[Proof of Proposition~\ref{prop:pareto}]
(i) follows from non-dominance; equal scores do not strictly dominate
one another. (ii) follows because the retained parent dominates the
candidate. For (iii), adding candidates cannot decrease the best
feasible accuracy over $V_n$. Removing a dominated point does not
change that maximum: a dominator has no higher cost and no lower
accuracy. Following dominators in a finite set reaches a non-dominated
point, so the maxima over $F_n$ and $V_n$ agree.
Fixed recorded scores are essential; this argument does not apply
to re-evaluations under changing policies.
\end{proof}

\section{Test-Leakage Audit}
\label{sec:leakage}

Because the proposer can edit harness code, it could insert hidden
\textsc{fail\_to\_pass} tests into the agent's context. We disable this
in the seed and check proposals for attempts to restore access
(\S\ref{sec:method:reward}). We examined two sets of runs.
In the main run, we found no hidden-test access in the code of the five
admitted harnesses. In a separate search with fixed policy weights,
the gate rejected all six proposals over six rounds for attempting to
restore test access. No candidate was admitted. The code inspection
does not rule out other information channels, and the six rejected
proposals do not estimate a general attack success rate or establish
that both safeguards are individually necessary. Training still uses
scalar task rewards.

\paragraph{Earlier evaluator pilot.}
An earlier 8B-policy pilot with a test/judge reward mixture exhibited
an increase in fabricated completion from 31\% to 78\% within 10 steps.
This pilot motivates evaluator-based verification but is distinct from
the binary-reward TB-2 experiments reported here.

\clearpage
\section{Hyperparameters}
\label{sec:appendix}

Table~\ref{tab:hparams} lists the SWE-rebench configuration.
Domain-specific settings are summarized in Table~\ref{tab:domain-protocols}.

\begin{table}[htbp]
  \centering
  \caption{Training configuration for \method{} on SWE-rebench.}
  \label{tab:hparams}
  \setlength{\tabcolsep}{4pt}
  \begin{tabular}{@{}lll@{}}
    \toprule

    \textbf{Group} & \textbf{Hyperparameter} &
    \textbf{Value} \\
    \midrule
    Model     & Policy                       & Qwen3.5-35B-A3B (MoE, $\approx$3B active) \\
              & Initialization               & Initial model checkpoint \\
              & Adaptation                   & LoRA, rank 32 \\
    \midrule

    RL        & Algorithm                    & GRPO, group-relative advantage \\

              & Importance-ratio clip (DAPO) & $[0.80,\, 1.28]$ \\

              & PPO epochs                   & 2 \\

              & Reference penalty / KL loss & Both disabled \\

              & Learning rate                & $2\times10^{-5}$ \\

              & Training steps         & 150 \\

              & Nominal task groups per step $B$ & 12 \\

              & Rollouts per group $K$       & 4 \\

              & Nominal rollouts per step ($BK$) & 48 \\
    \midrule
    Rollout   & Sampling temperature         & 0.7 \\
              & Episode timeout              & 600 s \\
              & Episode concurrency          & 8 \\
    \midrule

    Pool      & Capacity                     & 8 \\

              & Harness slots per step $N$ & 6 \\

              & Scheduler                    & difficulty-adjusted performance score, LCB guard \\
    \midrule
    Proposer  & Model                        & \texttt{claude-opus-4-7} \\
              & Trigger cadence              & Domain-, phase-, and pool-dependent \\
              & Harnesses per trigger        & up to 2 \\
              & SWE candidate checks         & contract $\to$ training-pool trials \\
    \midrule

    Evaluation & SWE strict resolution       & all \textsc{fail\_to\_pass} \& \textsc{pass\_to\_pass} \\

    \bottomrule
  \end{tabular}
\end{table}

\clearpage
\subsection{Domain-Specific Protocols}
\label{sec:protocols}

\paragraph{Training cost on SWE-rebench.}
In the 150-step SWE-rebench experiments, the number of training episodes
differs by less than 1\% between \method{} and RL-only. Harness validation adds
approximately 10\% overhead, with proposer inference incurring a separate
cost. These figures do not describe the separate 30-step ablation runs.
Similar training episode counts do not imply equal total compute.

\begin{table}[htbp]
\centering
\caption{Domain-specific and historical diagnostic configurations.
The last column describes earlier A1--A3 diagnostics, not the matched
30-step comparisons in Table~\ref{tab:ablation}. Sampling budgets are nominal.}
\label{tab:domain-protocols}
\begin{tabular}{@{}lrrr@{}}
\toprule
Setting & SWE & TB-2 & Historical A1--A3 \\
\midrule
Training entries & 400 & 843 & 400 \\
Task groups $\times$ rollouts/group & $12\times4$ & $18\times8$ & $8\times4$ \\
Training temperature & 0.7 & 0.9 & 0.7 \\
Episode timeout (s) & 600 & 1,200 & 600 \\
Policy-monitoring tasks & 32 & 20 & 32 \\
\bottomrule
\end{tabular}
\end{table}

\paragraph{Matched-budget ablation protocol.}
Table~\ref{tab:ablation} uses 30 training steps for the full method and
all five controls, with the same training budget as the full comparator
(12 task groups with four rollouts each per step). These runs are
separate from the 150-step main experiment. The full method and A2--A4
use eight harness-search attempts; A1 disables search. A4 performs its
eight attempts before RL, selecting the harness on 32 validation tasks
by strict resolved rate, with lower mean generated-token count breaking
ties. The selected harness (\texttt{candidate\_01}) is frozen during RL
and evaluated with the step-30 policy checkpoint. The full method,
A2, and A3 search during training. A5 searches at training steps
4, 8, 12, 16, 20, 24, and 28 (seven attempts) and evaluates the step-30
policy with the harness active at that step. A5 uses the same candidate checks
and acceptance rule as the full comparator, but keeps only one active
harness rather than a training population. All rows are evaluated on
the same 110 tasks with one attempt per task and strict resolution.
Training budgets are matched, but search counts are not identical across
all controls; candidate-evaluation and proposer costs can also differ.

\paragraph{SWE reward and filtering scores.}
\label{sec:reward-protocol}
For the full SWE run (B0), the pre-penalty training reward is
\begin{equation}
R_{\rm raw}=1.5(0.9f_{\rm F2P}+0.1f_{\rm P2P})
+\mathbf 1_{\rm Submitted}(0.05\,\mathrm{IF}
+0.05\,\mathrm{SBE}+0.075\,\mathrm{ETT}).
\end{equation}
Here $f_{\rm F2P}$ and $f_{\rm P2P}$ are the fractions of passing \textsc{fail\_to\_pass}
and \textsc{pass\_to\_pass} tests, respectively. IF is the fraction of
assistant turns containing exactly one recognizable shell code block;
SBE indicates a recognizable search or inspection operation before the
first edit; ETT indicates a recognizable test or validation command at
or after the first edit. SBE and ETT are binary indicators. Only the
process terms require the episode to end with \texttt{Submitted};
the correctness term is not gated by submission status.
For a normally retained sample, the reward used in the policy update is
\begin{equation}
R=R_{\rm raw}-P_{\rm len},\qquad
P_{\rm len}=0.1\operatorname{clip}
\left(\frac{L_{\rm seq}-(C-4096)}{4096},0,1\right),
\end{equation}
where $L_{\rm seq}$ is the prompt-plus-response training-sequence length
and $C$ is the training-sequence limit. This length differs from the
generated-token count $L$ in Eq.~\eqref{eq:grpo}. These reward weights
apply to B0, not uniformly to historical runs: the June configuration
used an F2P correctness weight of 0.85 rather than 0.9.

Filtering reads the harness-provided \texttt{test\_score}, not the
length-penalized reward $R$. For \texttt{swe\_mini\_v001}, this field is
the combined pass fraction over all target F2P and P2P tests, without
process rewards. Versions such as \texttt{swe\_v081\_clean} and
\texttt{swe\_v076} instead assign the harness-returned reward to this
field, so their filtering scores can contain process rewards but not
the subsequently applied length penalty. The retention diagnostic can therefore reflect variation in process
rewards as well as test correctness.

\paragraph{Training-batch preparation.}
Batch preparation applies overlength filtering before the group-variance
test. If fewer than two groups pass the variance test, the protective
fallback bypasses that test, not the preceding length filter. The
training adapter then removes groups with fewer than two surviving
sequences and aligns the remaining groups to a common size. A batch
with insufficient usable samples produces no policy update. The
diagnostic counting point precedes this adapter-level alignment
(Appendix~\ref{sec:retention}).

\paragraph{Candidate checks and policy monitoring.}
In the main SWE recipe, candidates enter population trials after contract
checks, and ranking is updated using actual SWE training rollouts.
A1 disables harness evolution; the matched comparison above uses common
candidate-check and acceptance rules for the full method and A5. TB-2 uses
contract checks, a one-task, one-episode smoke test, and candidate
validation on 20 tasks with one attempt per task at temperature zero,
before population trials. Candidate validation is distinct from
training-time policy monitoring: SWE monitoring uses 32 tasks under a
fixed baseline harness, while TB-2 monitoring uses 20 tasks. The
\texttt{ref\_check} monitoring routine runs every 10 training steps; it does not
imply an active reference-policy penalty.

\paragraph{Evaluation scope.}
The SWE main comparison evaluates 110 tasks with three attempts per
task and reports strict resolved avg@3; A1--A5
ablations use 110 tasks with one attempt per task.
SWE-bench Verified evaluates transfer without
further adaptation on 500 tasks with one attempt per task. TB-2 uses a
separately trained terminal agent. These benchmark evaluations are
distinct from candidate screening and training-time policy monitoring.
For our TB-2 evaluations, temperature is 0.6 and each task's binary
outcomes are averaged over the episodes actually returned; a task with no returned
episode contributes zero. The reported score is the arithmetic mean of
these task-level rates, rather than success in at least one attempt.
For a three-attempt evaluation, this equals fixed-denominator avg@3
only when all three episodes return for every task; missing episodes
on a partially completed task do not enter that task's denominator.
Four historical TB-2 evaluations each scheduled 89 tasks with three
attempts per task (267 episodes) at temperature 0.6; their returned
episode counts were 267, 255, 261, and 265, respectively.
The quoted Flat GRPO result instead follows Shepherd's original
89-task avg@5 protocol.

\paragraph{Scheduling and selection interfaces.}
The historical scheduler settings use proposal intervals of four steps
in phase A and eight in phase B, shortened to two steps in either phase
when the pool contains fewer than four members. The supplementary
A4/A5 comparisons follow the schedules specified above. A
proposal is eligible at positive scheduler steps divisible by that
interval, subject to the proposal budget. Saturation does not trigger
proposals. Forced-trial priority and minimum-use protection precede
ordinary eviction eligibility; archive dominance alone does not evict
a training-pool member. Policy-checkpoint selection and evaluation-harness
selection are performed separately. For the main SWE experiment, we use
the policy checkpoint at training step 150, paired with the highest-ranked
harness in the final training pool.

\section{Archival Training-Process Analysis}
\label{sec:archival}

\paragraph{Scope and counting protocol.}
Figure~\ref{fig:dynamics} uses scheduler steps 56--91 of the archived
\texttt{coevo\_8gpu\_sglang} segment, resumed from a checkpoint labeled
step~54. Each of the 36 steps has 48 logged training-rollout completions,
giving 1,728 completion events across 14 observed harness versions.
Shares use all 48 completions at the corresponding step as the
denominator. A version--step cell without a recorded completion is left
unassigned because pool membership is not established; it is not
interpreted as zero allocation for a confirmed pool member. Event
counts summarize records within this segment, not unique versions or
successful improvements. ``First completion'' denotes the first recorded
completion within this segment, not necessarily a version's first use.
The figure uses relative scheduler steps 1--36, corresponding to logged
steps 56--91. Labels H1--H14 correspond, in row order, to
\texttt{mini\_v001}, \texttt{v017}, \texttt{v024}, \texttt{v030},
\texttt{v032}, \texttt{v035}, \texttt{v036}, \texttt{v013},
\texttt{v037}, \texttt{v038}, \texttt{v039}, \texttt{v040},
\texttt{v041}, and \texttt{v042}, with the common \texttt{swe\_}
prefix omitted. These execution records do not determine which samples survived
filtering or contributed to a gradient update.

\paragraph{Development monitoring.}
The same archive reports development scores of 25\%, 18.75\%, 25\%, and
3.125\% at scheduler steps 60, 70, 80, and 90, respectively. All four
checks use the same logged baseline harness name and have 32 distinct
task-completion records. These 32 development tasks also occur in the
110-task set monitored during this run. We therefore treat this
sequence, including the decline at step~90, as development monitoring
rather than an independent-test curve. We do not infer a cause for the
decline or combine these points with a policy-only curve. The archival
segment is not used to reconstruct the main benchmark scores.

\paragraph{Configuration scope.}
The archived segment uses neither a reward-side reference penalty nor
a separate KL loss. Its launch settings specify a 75-turn episode limit
and a 32,768-token training-sequence limit; the latter is not a
serving-context limit. These settings apply to this segment rather
than uniformly to every experiment.
In this archived configuration, proposal intervals are four steps in
phase A and eight in phase B, shortened to two in either phase when
the pool contains fewer than four members. These are scheduler-step
intervals, not counts of successful optimizer updates.

\subsection{Training-Group Retention}
\label{sec:retention}
The following diagnostics come from historical runs, not the matched
30-step comparison in Table~\ref{tab:ablation}. In these diagnostic
configurations, the full run uses 12 nominal groups per step and A1--A3
use eight, all with four rollouts per group. These statistics describe
the earlier configurations only.
A group is a task--harness group within a logged scheduler step. Episodes are
deduplicated by \texttt{session\_id}; multiple response turns from one
episode do not count as separate rollouts. Let $\mathcal G_s$ denote
the groups after overlength filtering but before zero-variance
filtering, and let $K_{s,g}$ be the episode count in group $g$. Define
\begin{align}
I_{s,g}&=\mathbf{1}\!\left[K_{s,g}\ge2\;\land\;
\operatorname{Std}_{\mathrm{population}}
(z_{s,g,1},\ldots,z_{s,g,K_{s,g}})\ge10^{-6}\right],\\
F_s&=\frac{\sum_{g\in\mathcal G_s}I_{s,g}}{|\mathcal G_s|},
\qquad
\bar F=\frac{1}{|\mathcal S|}\sum_{s\in\mathcal S}F_s.
\end{align}
Here $z$ is the harness-specific SWE \texttt{test\_score}
(Appendix~\ref{sec:reward-protocol}), or
binary task reward for TB-2. For the SWE summaries below, $\mathcal S$
contains only steps with filtering records and nonempty group sets;
steps where the protective fallback skips filtering are excluded.
On these included steps, $F_s$ is the group-retention rate. The reported
mean $\bar F$ weights steps equally, rather than pooling groups across
steps. Multiple response turns never enter its denominator.

\begin{table}[htbp]
\centering
\caption{SWE group-retention diagnostics. Means are arithmetic averages
over recorded filtering steps, excluding protective fallback steps.
First-step values refer to the first included step, not necessarily the
first scheduler step.}
\label{tab:retention}
\begin{tabular}{@{}llr@{}}
\toprule
Configuration & Statistic & Group retention \\
\midrule
\method{} (full) & Mean over recorded filtering steps & 79.1\% \\
A1: Fixed-harness RL & Mean over recorded filtering steps & 52.2\% \\
\method{} (full) & First included step ($11/12$ groups) & 91.7\% \\
A1: Fixed-harness RL & First included step ($3/8$ groups) & 37.5\% \\
A2: No archive context & Mean over 27 recorded filtering steps & 64.2\% \\
A3: No execution traces & Mean over 24 recorded filtering steps & 74.0\% \\
\bottomrule
\end{tabular}
\end{table}

Table~\ref{tab:retention} summarizes these diagnostics.
The first included full-run step retains 11 of 12 groups (44 of 48
episodes at four episodes per group); A1 retains three of eight groups
(12 of 32 episodes). These are episode counts, not multi-turn response
counts. Because SWE filtering scores need not be binary and can include
process rewards in some harness versions, a group can pass the filter
without any fully resolved episode. Group retention can therefore
exceed the final resolved rate;
the binary-outcome bound in Proposition~\ref{prop:flywheel} does not
bound this statistic. The summaries characterize observed reward-score
variation, not gradient magnitude, convergence speed, or a causal
harness effect. They are separate from the utilization counts in
Figure~\ref{fig:dynamics}.

\subsection{Observational Harness-Edit Case}
\label{sec:case}

\begin{figure}[H]
\centering
\includegraphics[width=\linewidth]{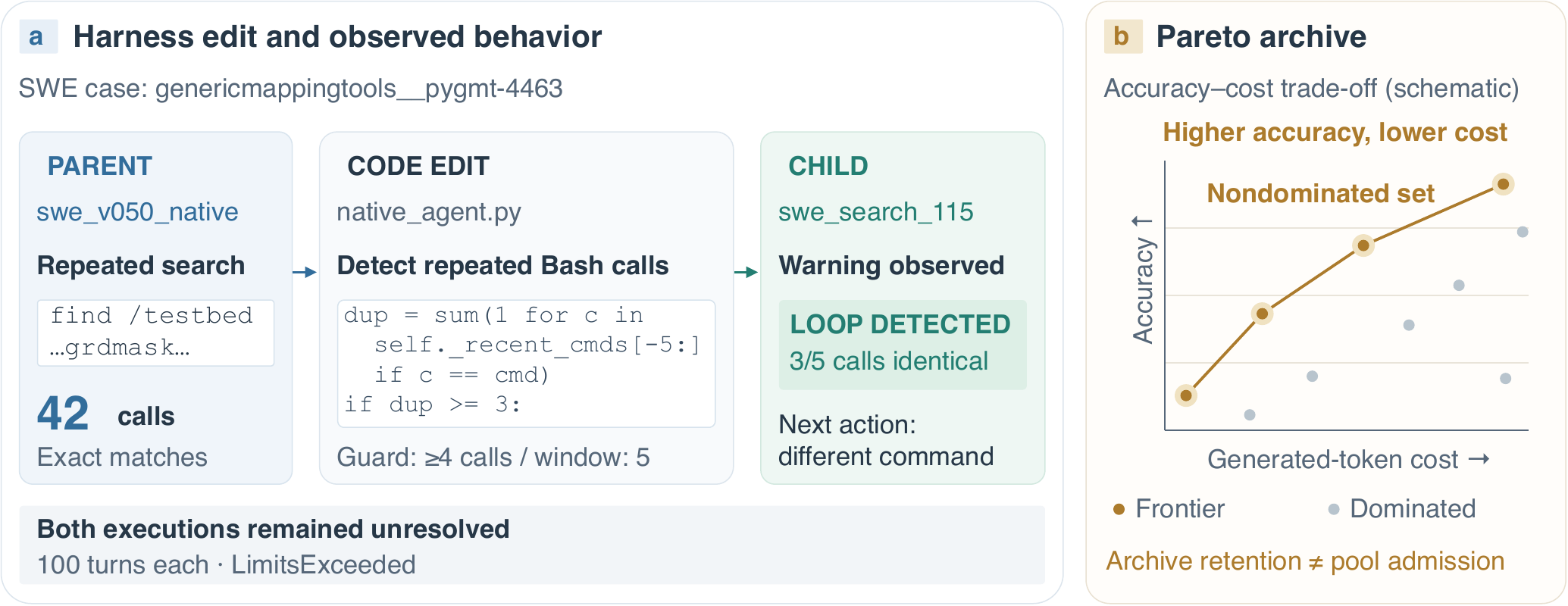}
\caption{Harness-edit example and historical Pareto archive.
(a) An observational parent--child example: a repeated-command warning
is followed by a different command, although both runs remain unresolved.
Code excerpts are abridged. (b) A separate schematic of accuracy--cost
archive retention, not measured results from the example in (a).
Archive retention is distinct from training-pool admission.}
\label{fig:evolution}
\end{figure}

Figure~\ref{fig:evolution}(a) uses task
\texttt{genericmappingtools\_\_pygmt-4463} from an independent SWE harness
search. The parent, \texttt{swe\_v050\_native}, makes 42 identical calls
to a search command. The child, \texttt{swe\_search\_115}, adds repeated-command
detection in \texttt{native\_agent.py}, inside \texttt{get\_observation()}.
The displayed source excerpt counts matches within the last five calls,
with a guard requiring at least four recorded calls and a warning
threshold of three matches. The child trace contains the warning
``LOOP DETECTED (3/5 recent calls identical)'', followed by a different
Bash command. Both executions end at 100 turns with
\texttt{LimitsExceeded}, leaving the task unresolved.
These are observational traces, not controlled replays at a fixed policy
checkpoint; admission of this child to the main training pool is not
established. The traces show a change in commands after the warning, but no task-level
improvement.
Panel (b) is a separate conceptual depiction of the historical Pareto
archive. Its point positions are schematic and are not measurements
of the parent and child in panel (a).

\section{Limitations}
\label{sec:limitations}
We evaluate one policy-model setting using single training runs,
without confidence intervals or a complete comparison of total compute.
The reward-contrast analysis does not prove faster convergence, and
final task counts cannot establish that gradient instability caused
the single-harness control's lower performance. Repeated validation can
overfit harness selection; policy updates can also make archive scores
stale. Fixing the action-format family restricts which edits can be
explored. We do not separately measure the effects of proposer quality
or policy gains under a common harness. Pretraining contamination remains
possible in the cross-dataset evaluation. The terminal results use a
separately trained agent, not zero-shot transfer from SWE.

\clearpage
\section{Comparison with Closely Related Frameworks}
\label{sec:comparison}

Table~\ref{tab:framework-comparison} compares adaptation targets and
training organization. Joint adaptation is shared with prior work;
\method{} maintains a population that collects training rollouts for one
shared policy, rather than only a search archive or phase-local candidate pool.

\begin{table}[htbp]
\centering
\caption{Design comparison with closely related frameworks, based on
their described methods. These distinctions are not a performance ranking.}
\label{tab:framework-comparison}
\setlength{\tabcolsep}{4pt}
\begin{tabular}{@{}>{\raggedright\arraybackslash}p{0.16\linewidth}>{\raggedright\arraybackslash}p{0.25\linewidth}>{\raggedright\arraybackslash}p{0.25\linewidth}>{\raggedright\arraybackslash}p{0.27\linewidth}@{}}
\toprule
\textbf{Framework} & \textbf{Adaptation target} & \textbf{Edit generation} & \textbf{Training organization} \\
\midrule
EvoTrainer\newline\citep{evotrainer} & Training-side diagnostics and training recipes
& Trainer agent diagnoses artifacts and proposes interventions
& Versioned training branches with promotion, pruning, and skill reuse \\
\addlinespace
WHALE\newline\citep{whale} & Executable harness and model weights
& Meta-Harness proposer searches under frozen weights
& Weight updates under one selected harness alternate with search; instantiated with online RSFT \\
\addlinespace
HASE\newline\citep{hase} & Guidance harness; local evaluator where permitted
& Task policy emits both solutions and harness edits
& GRPO with rollout-local edits; reviewed candidates update one persistent harness at phase boundaries \\
\addlinespace
HarnessForge\newline\citep{harnessforge} & Population of harness--policy pairs
& Meta-agent uses trajectory feedback and an archive
& Each surviving harness receives a dedicated policy adapter \\
\addlinespace
HarnessBandit\newline\citep{harnessbandit} & Shared policy and harness scheduling
& No online harness-code evolution
& One harness per update, selected from a given set using learnability and transferability signals \\
\addlinespace
\method{} & Population of executable agent-side harnesses and shared policy weights
& Separate proposer uses current-policy traces and a historical Pareto archive
& Shared-policy RL across scheduled pool members; accepted candidates enter at step boundaries \\
\bottomrule
\end{tabular}
\end{table}

WHALE permits replacing both update operators, so using GRPO rather than
its RSFT instantiation is not, by itself, the defining distinction.
HASE has a phase-local candidate pool and preserves an immutable external
evaluator even when a local proxy evaluator can be repaired. Our method
instead separates code proposal from task-policy actions and fixes the
task evaluator. This comparison describes architectural differences, not a performance ranking.

The fixed-harness control disables both evolution and population
diversity, so it does not isolate the benefit of population-based training.
Search-then-RL (A4) resolves 25/110 tasks versus 28/110 for \method{} under
matched training budgets (Table~\ref{tab:ablation}), but tests one-time
search, not WHALE's repeated alternation. The single-harness alternating
control (A5) resolves 21/110 tasks with a matched training budget,
but is a local control rather than a
reproduction of WHALE. Comparisons require aligned update and search
budgets; these results do not
establish superiority over the frameworks in Table~\ref{tab:framework-comparison}.

\section{Descriptive Interaction of Reported Gains}
\label{sec:interaction}
For reported scores $S$, define
\begin{equation}
I=S_{\rm joint}-S_{\rm harness}-S_{\rm RL}+S_{\rm base}.
\label{eq:interaction}
\end{equation}
The values are 3.4 percentage points for SWE-rebench, 2.2 for SWE-bench
Verified, and $-4.5$ for TB-2. On SWE-rebench, the joint gain is
$10.3/(3.9+3.0)=1.49$ times the sum of the separate gains. The negative
TB-2 contrast combines our evaluations with the quoted Shepherd baseline
and is not protocol-matched. These contrasts summarize the reported scores; they do not estimate
causal interactions under matched budgets or provide statistical tests.

\end{document}